\documentclass[runningheads]{llncs}
\usepackage[T1]{fontenc}
\usepackage{booktabs}
\usepackage{graphicx}
\usepackage{array} % For defining new column types
\usepackage{subcaption}
\usepackage{hyperref}
\usepackage{color}

\begin{document}
\title{Automating 3D Dataset Generation with Neural Radiance Fields}
\author{Paul Schulz\inst{1}\orcidID{0009-0006-3444-5800} \and
Thorsten Hempel\inst{1}\orcidID{0000-0002-3621-7194} \and
Ayoub Al-Hamadi\inst{1}\orcidID{0000-0002-3632-2402}}

\authorrunning{P. Schulz et al.}
%\authorrunning{X et al.}
\institute{Otto von Guericke University, Universitätsplatz 1, 39104 Magdeburg, Germany}
\maketitle{THIS WORK HAS BEEN SUBMITTED TO SPRINGER FOR PUBLICATION. COPYRIGHT MAY BE TRANSFERRED WITHOUT NOTICE, AFTER WHICH THIS VERSION WILL NO LONGER BE ACCESSIBLE.}         % typeset the header of the contribution
\begin{abstract}

3D detection is a critical task to understand spatial characteristics of the environment and is used in a variety of applications including robotics, augmented reality, and image retrieval. %th 
Training performant detection models require diverse, precisely annotated, and large scale datasets that involve complex and expensive creation processes. Hence, there are only few public 3D datasets that are additionally limited in their range of classes.
In this work, we propose a pipeline for automatic generation of 3D datasets for arbitrary objects. %th 
By utilizing the universal 3D representation and rendering capabilities of Radiance Fields, our 
pipeline generates high quality 3D models for arbitrary objects. These 3D models serve as input for a synthetic dataset generator.
Our pipeline is fast, easy to use and has a high degree of automation. Our experiments demonstrate, that 3D pose estimation networks, trained with our generated datasets, archive strong performance in typical application scenarios.

\end{abstract}

\begin{figure}[h]
	\centering
		\includegraphics[width=1\textwidth, clip]{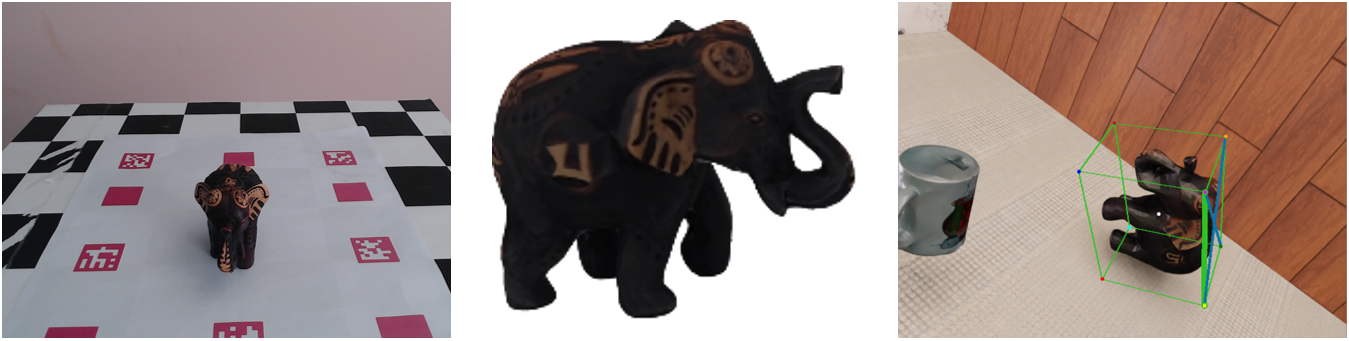}
	\caption{Automated dataset generation with our pipeline. The process start with capturing images of an object, continues with 3D model creation and finishes with 3D dataset generation.}
	\label{fig:teaser}
\end{figure}

\section{Introduction}

3D scene perception and manipulation is a fundamental task in various fields where autonomous agents interact with their environment. 
Applications in autonomous driving, mobile robotics, medicine or manufacturing~\cite{CARLA,SLAM,Bin_Pick_large_scale,Surgical} and intuitive human-robot interaction systems~\cite{RoSA_1,RoSA_2} rely on this capability.

To effectively perform scene manipulation, an agent has to estimate  the pose of relevant objects within its environment.
State-of-the-art pose estimation methods utilize neural networks that require large, diverse and precisely annotated 3D datasets for training. 
However, generating this data is a significant challenge, because collecting and annotating 3D datasets is labor-intensive and expensive, particularly when tailored datasets are required for specific applications.

Recent works proposed approaches to automate 3D dataset generation, by incorporating 3D models of the objects of interest~\cite{BOP,Fallings_things,Bin_Pick_Sim_Real,Bin_Pick_large_scale,Homebrew,YCB,Phocal}.
These methods either generate real or synthetic datasets. Automating real world dataset generation requires dedicated equipment and is limited in its field of application~\cite{Homebrew,YCB,Phocal}. Synthetic dataset generation does not have these limitations, because it only requires 3D modeling software and the workflow can be easily adapted~\cite{BOP,Fallings_things,Bin_Pick_Sim_Real,Bin_Pick_large_scale}.

However, real-world and synthetic approaches assume that 3D models of the target objects are available. Creating these models with traditional methods, like 3D scanning or 3D modeling, is expensive and requires expert knowledge. Generating even a single high quality model for an individual target object may require, a textureless CAD model 3D scanning equipment and model refinement~\cite{MVML}

We address these limitations by proposing a pipeline, which incorporates 3D model creation into the dataset generation process.
To archive this, our pipeline creates 3D models from 2D images of target objects by utilizing Radiance Fields. 
Afterward, our pipeline incorporates these models in a state-of-the-art dataset generation workflow.

In contrast to previous works, our pipeline generates datasets for the pose estimation of arbitrary objects without requiring 3D models.
To verify our approach, we applied our pipeline to six objects of varying complexity and trained a 3D pose estimation model on our generated datasets. Furthermore, we tested the trained networks for two tasks: in-hand pose estimation and tabletop pose estimation. 
In summary, our contribution is an end-to-end dataset generation pipeline, which begins with arbitrary objects and ends with diverse, precisely annotated 3D datasets for these objects.

\section{Related Works}

To the best of our knowledge, our method is the first to automate dataset generation by utilizing Radiance Fields for 3D model creation. Most similarly, Wong et al. \cite{Syn_DS} introduced a synthetic dataset generation pipeline, but in contrast to us, they use photogrammetry to create 3D models from 2D images. In the following, we provide an overview of synthetic dataset generation methods and Radiance Field frameworks that could potentially be integrated into our pipeline.

\subsection{Automated Generation of Synthetic 3D Datasets}

\subsubsection{Dataset Generation Methods}
Synthetic dataset generation takes place in virtual 3D environments, like game engines~\cite{Fallings_things} or general 3D modeling software~\cite{BOP}.
The dataset generation workflow, proposed by state-of-the-art publications, can be divided in two phases: scene composition and rendering + annotation. Scene composition constructs 3D scenes by placing objects in a virtual environment. These objects include target objects and additional content like distractors. The subsequent rendering + annotation process generates RGB images from the scenes, by applying 3D rendering software. Since all parameters of the 3D scene are known, the step also performs automated annotation~\cite{Fallings_things,Syn_DS,Bin_Pick_large_scale}.

Recent works can be divided in general and specific generation methods. General methods can be applied to arbitrary domains, specific method incorporate domain restrictions and are therefore applied in specific domains.

Hodan et al.~\cite{BOP} created a general generator for the BOP challenge. Tremblay et al.~\cite{Fallings_things} developed another general method for the pose estimation of arbitrary household objects.
Both methods are suitable for arbitrary target objects, because they do not incorporate domain restrictions. In contrast to that, 
Sun et al.~\cite{SHIFT} developed a specific method, which creates datasets for the autonomous driving domain. Their framework incorporates special domain shifts for autonomous driving, like changing weather conditions. 
Other specific methods~\cite{Bin_Pick_Sim_Real,Bin_Pick_large_scale} tackle industrial tasks, like bin packing. These methods apply domain restrictions in the form of special industrial environment. Furthermore, McCormac et al.~\cite{Scene_Net_RGB_D} applied a specific dataset generator to create a benchmark dataset for household scenes.

\subsubsection{Dataset Generation Frameworks}
Recent works also introduced dedicated frameworks to develop customized dataset generators~\cite{Blenderproc,Kubric,Replicator}.
These frameworks incorporate measurements to approach a central limitation of synthetic dataset generation, which is known as the 
simulation to reality gap (Sim2Real gap). 
The Sim2Real gap causes a performance drop, when networks trained on synthetic data are applied in the real world. 
The gap can be divided into the appearance gap and the content gap.
The appearance gap is caused by differences in detailed, pixel-level scene appearances, resulting from a lack of realism~\cite{ISAAC,Meta_Sim}. The content gap is related to the context of simulated scenes. It appears when the scene content does not reflect the variety of objects and environments found in the real world~\cite{ISAAC,Meta_Sim,Structured_Domain_Randomization}.
Dataset generation frameworks incorporate photorealistic rendering and domain randomization to reduce the Sim2Real gap. Photorealistic rendering reduces the appearance gap by making objects and environments appear more realistic. 
Domain randomization reduces the content gap by randomizing the texture of objects of interest and/or the 3D environment~\cite{Blenderproc,Replicator}.

\subsection{Radiance Fields}
In 2020, Mildenhall et al.~\cite{Nerf} introduced Neural Radiance Fields (NeRF) for novel view synthesis. 
NeRFs combine the function approximation capability of multi layer perceptrons (MLP) with principles of volume rendering.
NeRFs learn an unrestricted volumetric 3D representation and the view dependent appearance of an object or a scene from 2D images. Furthermore, they utilize volume rendering principles like ray marching to render photorealistic views from this representation\cite{Nerf,Advances}.

Dataset generators require 3D models of the object in the form of meshes or pointclouds. Extracting such representations is straight forward: In the first step one applies a volume sampling algorithm like Marching Cubes to transfer the volume in a mesh, afterward, one projects the color of the images onto the mesh surface. However, applying this approach, leads to a significant fidelity drop. 
Despite a tendency to learn smooth volumes~\cite{Nerf_plus_plus}, NeRFs learn unrestricted volumes. Therefore, the surfaces extracted from the volume may not resemble the actual surface of the scene. \cite{Delicate,NeuS,BakedSDF}
We will refer to this problem as unconstrained surface issue. 
Recent works approach this problem by regularizing NeRF to resemble surfaces~\cite{Unisurf,VolSDF,NeuS}. 
Wang et al.~\cite{NeuS} and Yariv et al.~\cite{VolSDF} approximate the zero level set of the SDF with volume rendering. Oechsle et al.~\cite{Unisurf} unify volume and surface rendering by learning an occupancy field. 3D models exported from these representations have a more accurate surface than 3D model exported from standard NeRF models. 

NeRFs model view dependent effects as part of the appearance. With the goal of high quality mesh export in mind, this feature can lead to difficulties, since view dependency might be baked into the mesh texture. A way to migrate this problem, is to decompose the appearance in reflectance and illumination~\cite{NeRD}. 

Tang et al.~\cite{Delicate} present a pipeline for textured mesh generation, which addresses this weakness. 
They optimize a NeRF and utilize two MLPs for color rendering to separate view-dependent colors into diffuse and specular components. Afterward, they apply a mesh refinement strategy, which adjusts mesh vertices and faces. The diffuse color is baked in a diffuse texture. Combined with the refined mesh, it can be used in traditional 3D modeling software.

\section{Methodical Approach}
\begin{figure}[t]
	\centering
		\includegraphics[width=1\textwidth, clip, trim={0.7cm 0 0.7cm 0}]{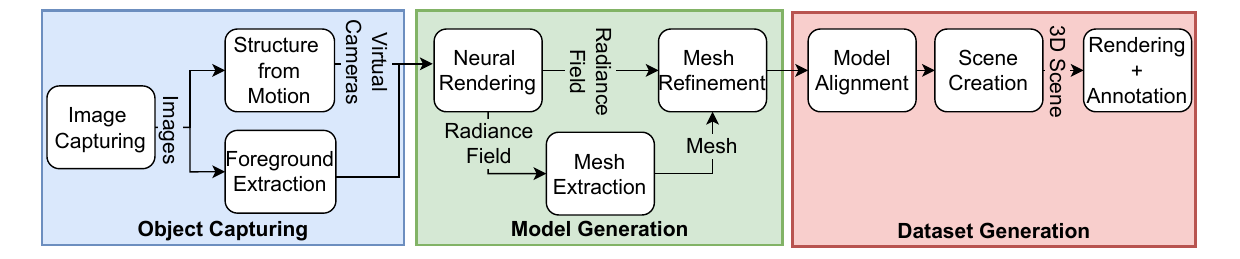}
	\caption{Automated dataset generation pipeline, each block is depicted with its corresponding phases.}
	\label{fig:_automated_dataset_generation_pipeline}
\end{figure}

Our goal was to create a reliable, fast and easily adaptable synthetic dataset generation pipeline for 3D Pose Estimation. The pipeline should work 
for arbitrary objects, without requiring 3D models of the target objects.
We propose a pipeline that expands a state-of-the-art synthetic dataset generator with a neural rendering framework, specialized on 3D model creation. \footnote{The implementation of our pipeline is available under: \url{https://github.com/PaulSK98/Nerf2Dataset.git}} The design of this pipeline is illustrated in Fig.~\ref{fig:_automated_dataset_generation_pipeline}.
In the following, we will introduce each component of the pipeline in detail.

\begin{figure}[t]
	\centering
		\includegraphics[width=1\textwidth, clip]{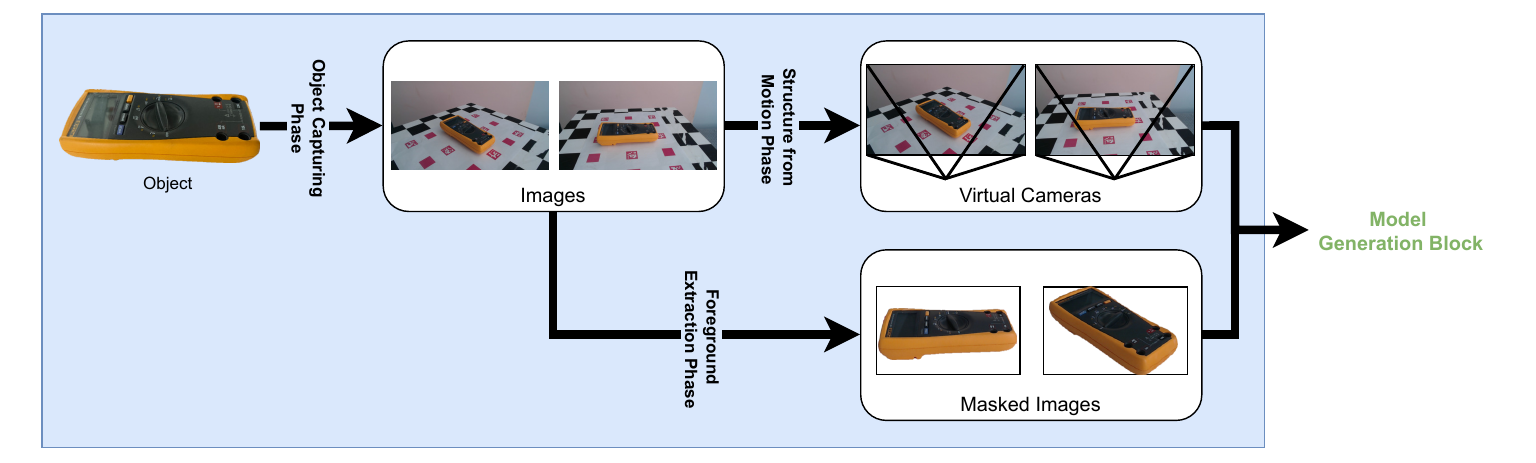}
	\caption{Design of our Object Capturing block, each capturing phase is displayed with its in- and output. The process begins with image capturing. Structure from Motion and Foreground Extraction perform post-processing on the captured images.  }
	\label{fig:Object_Capturing_Block}
\end{figure}

\subsection{Object Capturing}
Object Capturing is the first block of our pipeline. It captures 2D images of a target object and prepares them to be used to train a Radiance Field.
The design of our Object Capturing process is depicted in Fig.~\ref{fig:Object_Capturing_Block}. 

\begin{figure}[t]
	\centering
		\includegraphics[width=0.6\textwidth, clip]{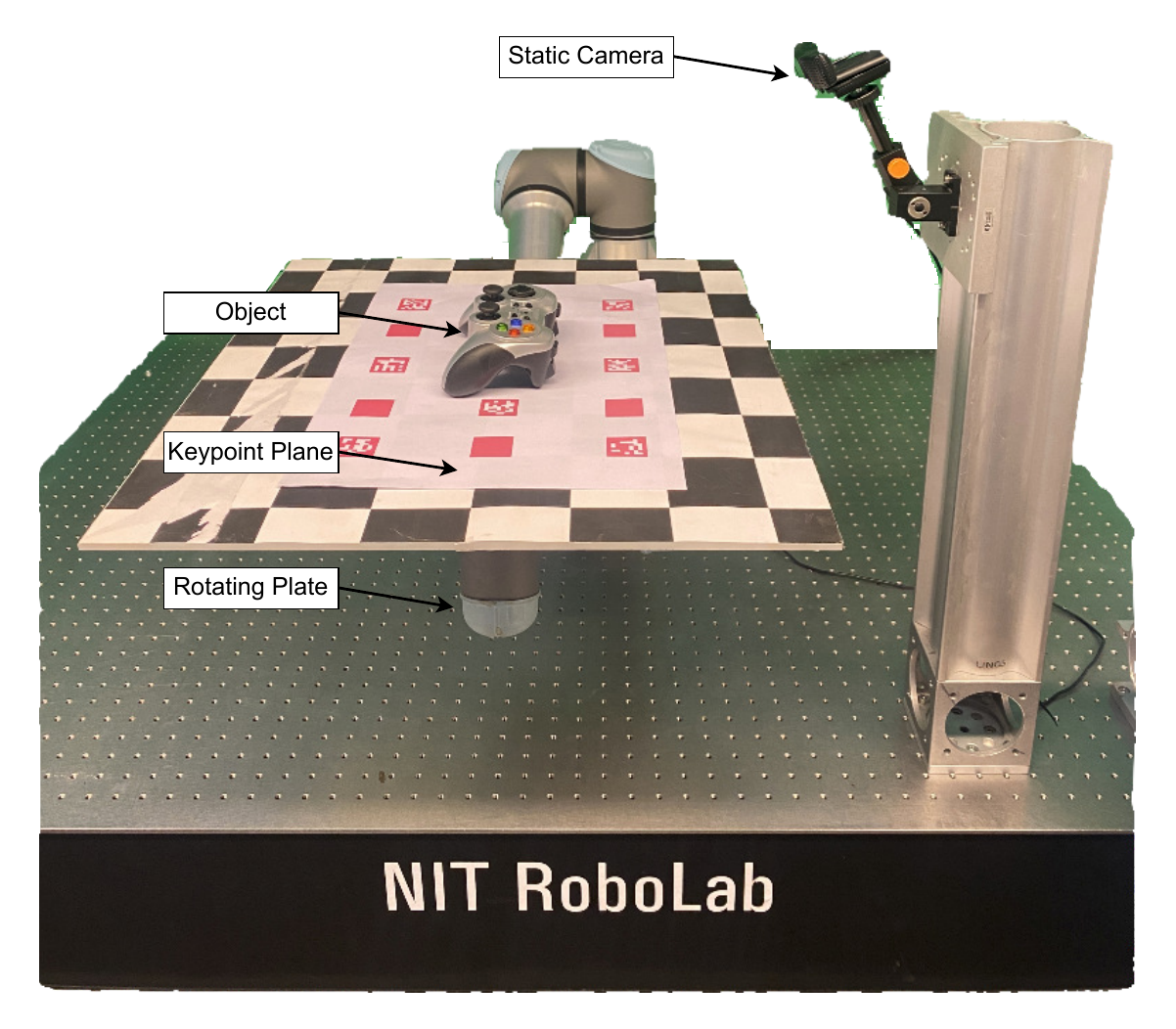}
	\caption{Object capturing set up, we placed the object on a rotating plate and captured it with a static camera.}
	\label{fig:Object_Capturing_set_up}
\end{figure}

The process begins with capturing images of the object of interest. Our setup to perform this task is shown in Fig.~\ref{fig:Object_Capturing_set_up}.
Radiance Fields work well when maintaining a constant camera-to-object distance, varying distances may lead to rendering artifacts~\cite{Mip_Nerf}. Additionally, Object Capturing should include many perspectives to prevent overfitting to specific views. To ensure both aspects, we captured the object in a 360-degree circular trajectory, with the object placed on a rotating plate and the camera kept static. We masked the static scene parts of the background during this process. We chose this setup because it is easily reproducible by others, as it only requires a fixed camera and a plate to rotate the object. % It is also easy to automate, which further aligns with our research goals. 
Radiance Fields assume a static object of interest, a static light source and a moving camera. To reduce additional view-depended effects which might be introduced by hurting this assumption, we use diffuse background light during capturing. 

Applying Structure-from-Motion (SfM) is the first of two post-processing steps after image capturing. 
Neural rendering frameworks utilize SfM pipelines like COLMAP~\cite{COLMAP} to estimate camera poses for real-world images. Since COL-MAP is the de-facto standard for this step, we applied COLMAP as well. COLMAP relies on a sufficient number of 2D features and 2D-to-3D correspondences to estimate camera positions. To support this process for objects with fewer features, we used a plane with tags as the background during object capturing.
Foreground extraction is the second post-processing step, applied during Object Capturing. It ensures that the Radiance Field learns the target object rather than the background. We applied an off-the-shelf foreground extractor to perform this task.

\begin{figure}[t]
	\centering
		\includegraphics[width=1\textwidth, clip]{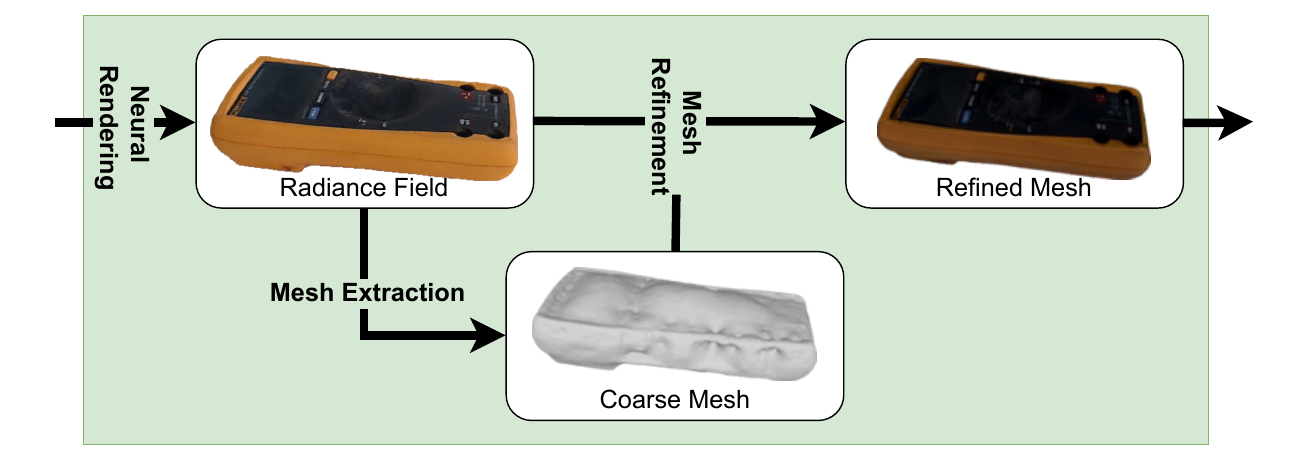}
	\caption{Design of our Model Generation block, it receives the output of the previous block as input and generates a textured 3D model.}
	\label{fig:Model_Generation_Block}
\end{figure}
\subsection{Model Generation}
Model Generation is the second stage of our pipeline. It takes the output from the previous block (virtual cameras and masked images) as input and creates high-quality textured meshes.
To achieve this, we integrated a NeuS implementation~\footnote{The implementation is available under: \url{https://github.com/ashawkey/nerf2mesh}} of the Nerf2Mesh framework by Hoang et al.~\cite{Delicate} into the design of the Model Generation Block. The block design is displayed in Fig.~\ref{fig:Model_Generation_Block}
As mentioned in the previous section, Radiance Fields struggle with poor texture mapping. Volumetric representations additionally suffer from the unconstrained surface issue. The NeuS-based Nerf2Mesh implementation migrates both of these issues by incorporating a neural surface and by separating the texture into a diffuse and a specular branch.
Neural Rendering is the first step in the Model Generation process. It trains a NeuS-based Radiance Field using the virtual cameras and masked images, provided by the Object Capturing process.
Furthermore, Neural Rendering applies a Multiresolution Hash Encoding~\cite{InstantNGP} to speed up the training process. 
Afterward, the Marching Cubes algorithm extracts a coarse mesh from the Radiance Field. In the last step of Model Generation, Mesh Refinement optimizes the mesh geometry and appearance through differentiable rendering and non-differentiable face adjustment. In the end, a textured mesh is exported, by mapping the texture on the refined mesh.

\subsection{Dataset Generation}
Dataset Generation is the third and final step of our pipeline. It creates a synthetic 3D dataset for 6D Pose Estimation of the target object. Our goal is to design a dataset generation pipeline for arbitrary objects. Therefore, we implemented a general dataset generator. Because dataset generation frameworks integrate measures, which approach the Sim2Real gap, we implemented the block via BlenderProc~\cite{Blenderproc}.

\begin{figure}[t]
\centering
\includegraphics[width=1\textwidth]{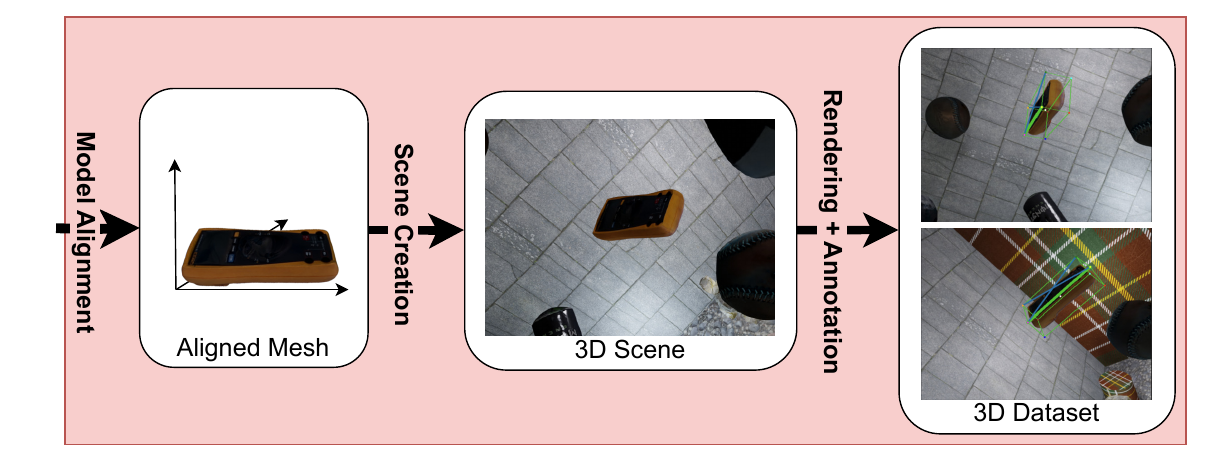}
\caption{Design of our Dataset Generation block, based on the textured meshes of the subsequent block, it generates a 3D Dataset in automated fashion.}
\label{fig:dataset_generation_block}
\end{figure}

Fig.~\ref{fig:dataset_generation_block} illustrates our dataset generation workflow. Model Alignment is the first phase of this process. It aligns the textured meshes generated in the previous block with the coordinate axes of the world coordinate system.
This step is required, to determine the 3D position of the target objects after Scene Composition. In the next step, Scene Composition creates domain randomized 3D scenes which contain light sources, virtual cameras, the objects of interest and distractors.
Lastly, a Rendering + Annotation phase renders images from the virtual cameras and annotates the target objects present in the image. It utilizes BlenderProc's photorealistic rendering capabilities to approach the appearance gap.

\section{Experiments}

%\subsection{Experimental Setup} 

\newcolumntype{C}{>{\centering\arraybackslash} m{0.25\textwidth}}

\begin{table}[t]
    \centering
    \begin{tabular}{C C C}
        % Row 1 - Simple
        \textbf{Simple} &
        \includegraphics[width=0.1\textwidth]{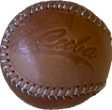} &
        \includegraphics[width=0.18\textwidth]{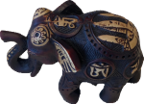} \\
        & Baseball & Elephant \\
        % Row 2 - Intermediate
        \textbf{Intermediate} &
        \includegraphics[width=0.12\textwidth]{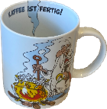} &
        \includegraphics[width=0.2\textwidth]{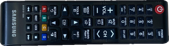} \\
        & Cup & Remote \\
        % Row 3 - Complex
        \textbf{Complex} &
        \includegraphics[width=0.18\textwidth]{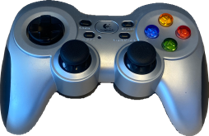} &
        \includegraphics[width=0.25\textwidth]{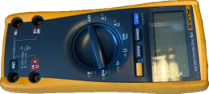} \\
        & Controller & Multimeter \\
    \end{tabular}
    \caption{Target objects, categorized according to complexity.}
    \label{tab:objects_of_interest_categorized}
\end{table}

We tested Object Capturing and Model Generation for all these objects. Dataset Generation was tested for one object per category. All experiments were conducted with an NVIDIA Quadro RTX 8000 graphics card.

% Definea new column type for centering content
\newcolumntype{C}{>{\centering\arraybackslash} m{0.3\textwidth}}

% Begin the table environment
\begin{table}[t]
    \centering
    \begin{tabular}{C C C}
        \hline
        % Column headers (centered)
        \textbf{Image Capturing} & \textbf{Structure from Motion} & \textbf{Foreground Extraction} \\
        \hline
        % Row 1 - Elephant
        \includegraphics[width=0.28\textwidth]{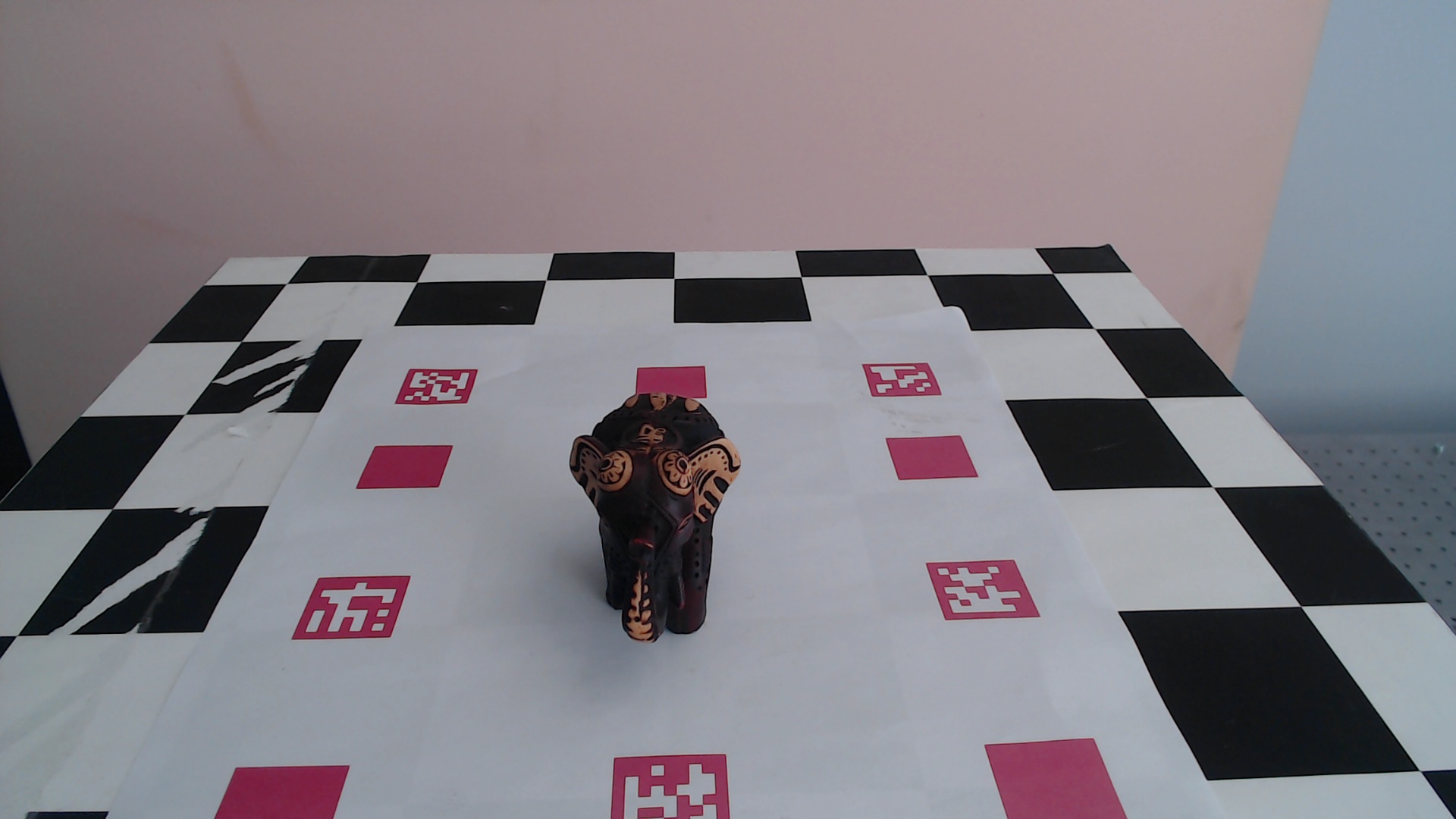} &
        \includegraphics[width=0.22\textwidth]{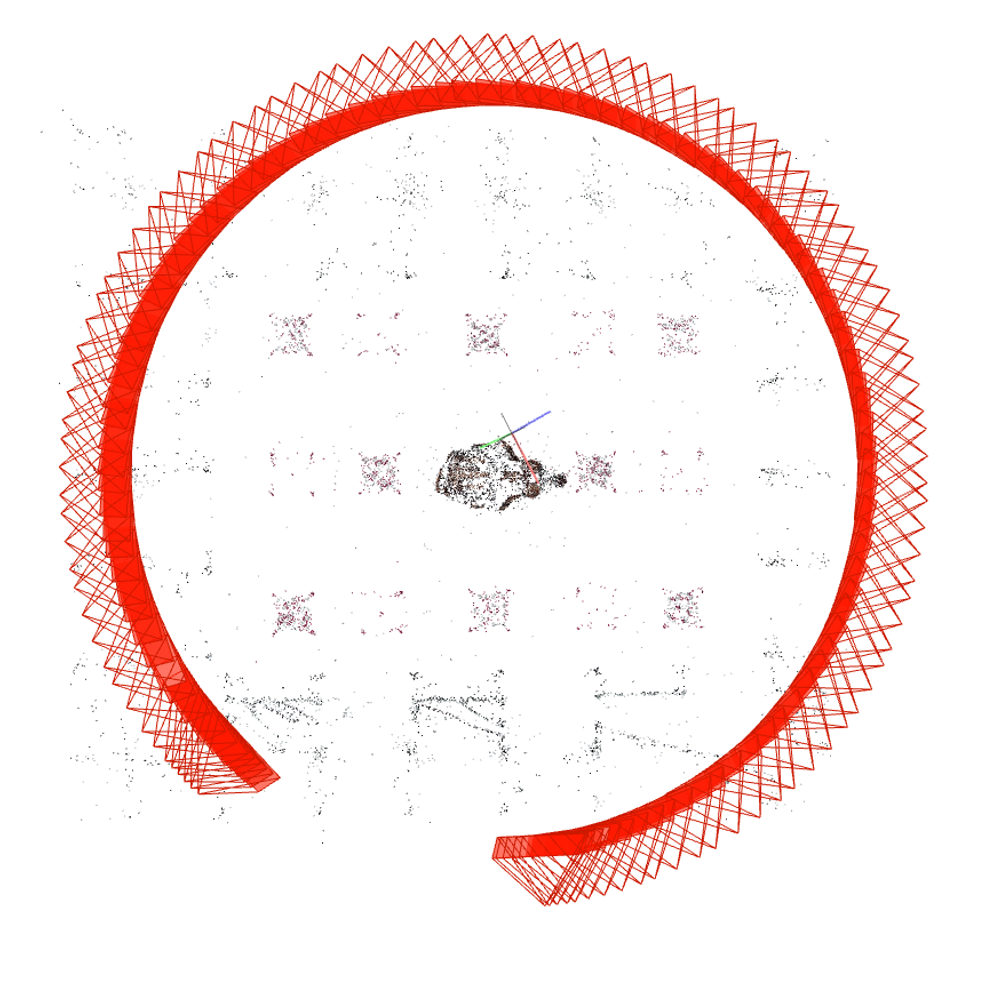} &
        \includegraphics[width=0.12\textwidth]{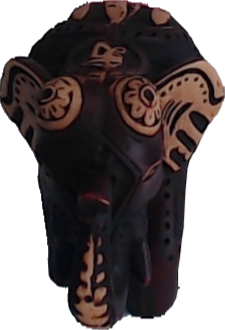} \\
        % Row 2 - Remote
        \includegraphics[width=0.28\textwidth]{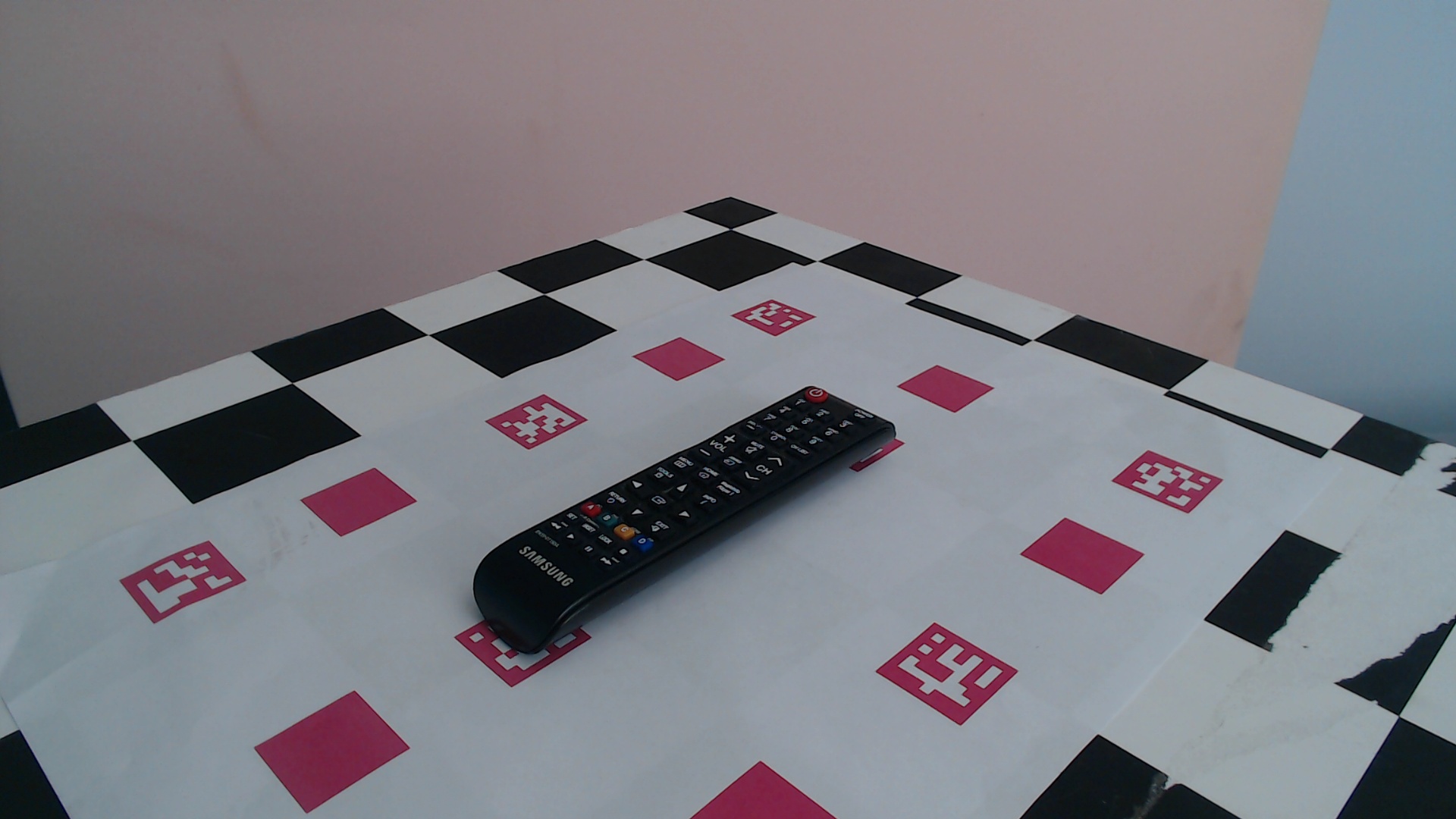} &
        \includegraphics[width=0.225\textwidth]{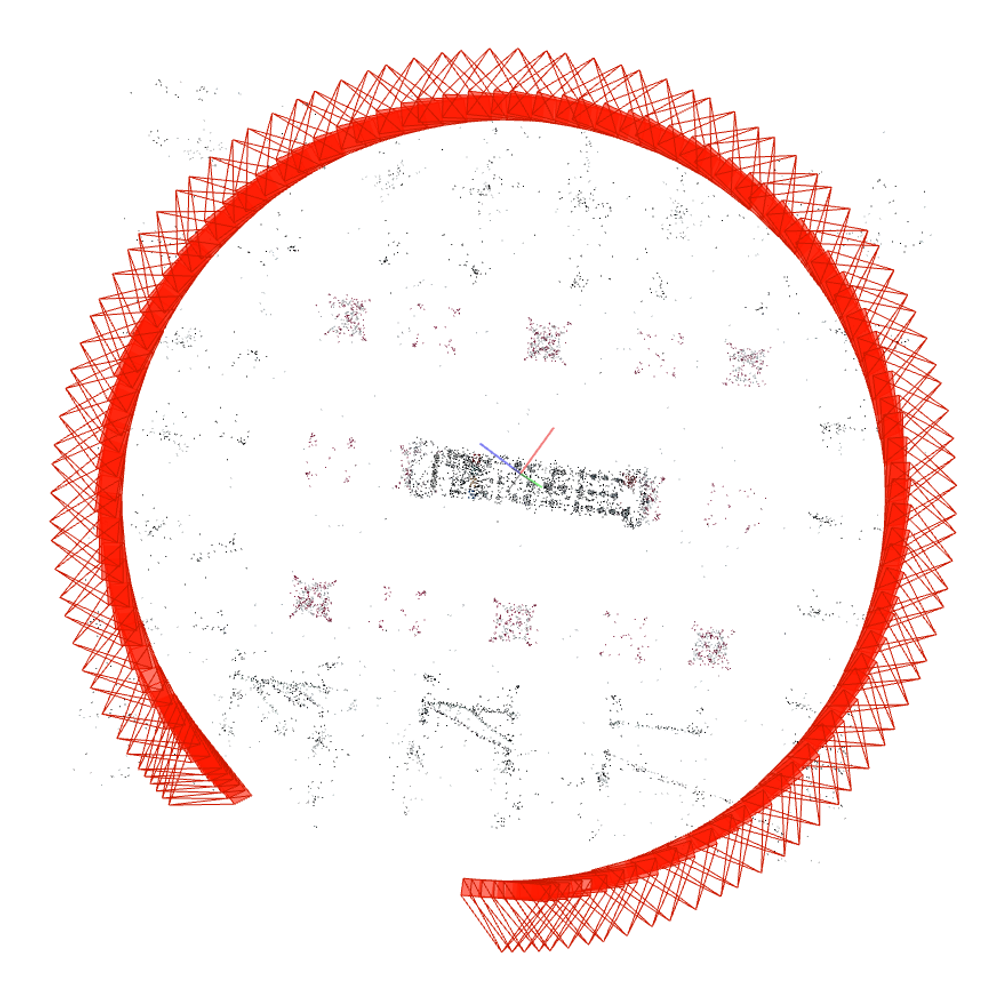} &
        \includegraphics[width=0.25\textwidth]{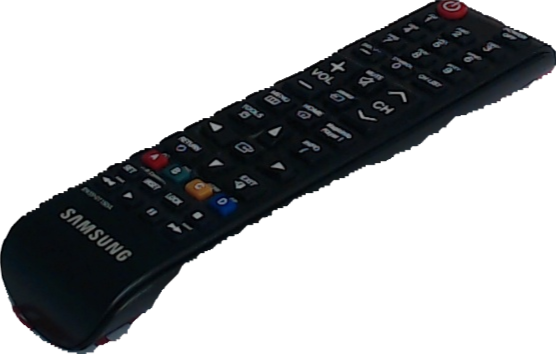} \\
        % Row 3 - Multimeter
        \includegraphics[width=0.28\textwidth]{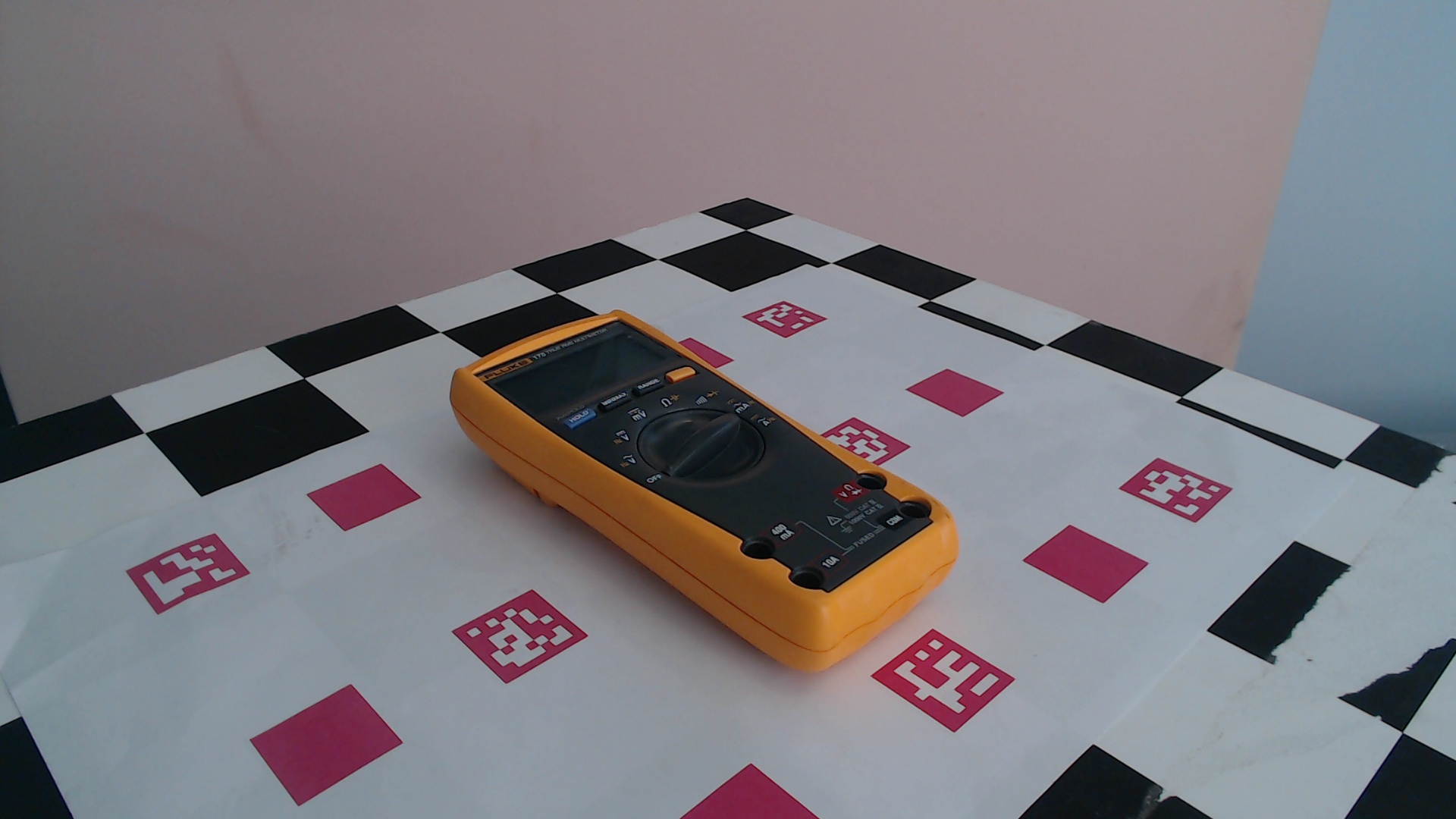} &
        \includegraphics[width=0.22\textwidth]{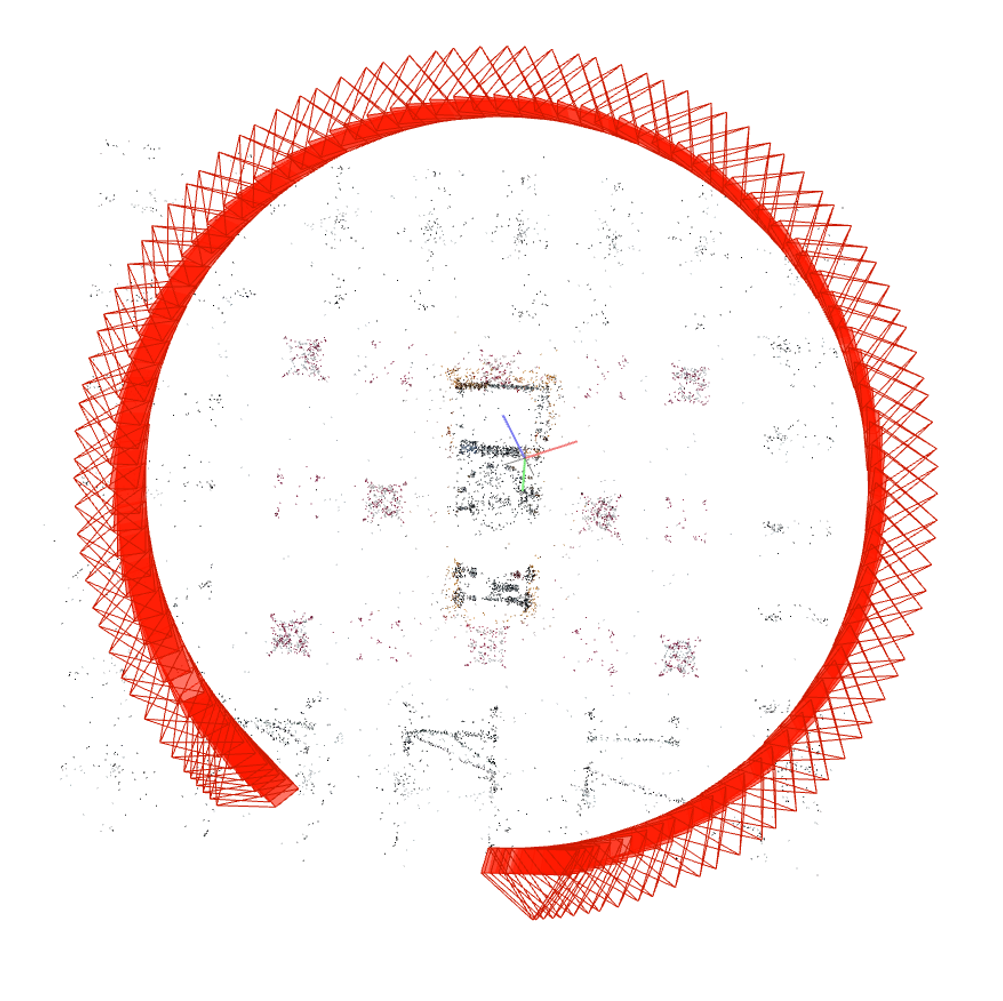} &
        \includegraphics[width=0.22\textwidth]{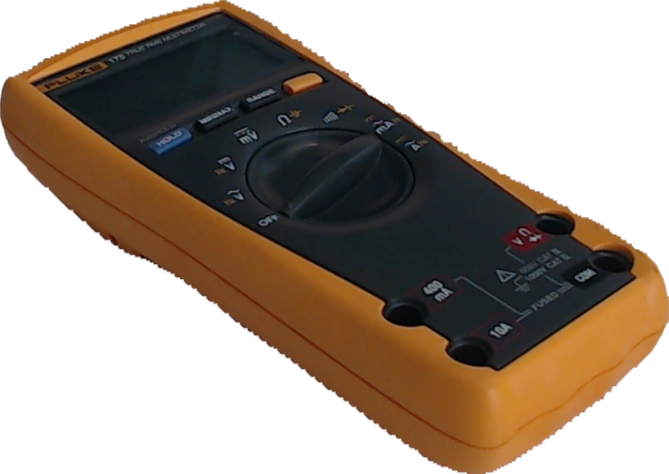} \\
        \hline
    \end{tabular}
    \caption{Object Capturing for three target objects (elephant, remote, multimeter).}
    \label{Qualitative_Results_Object_Capturing_Block}
\end{table}
\subsection{Object Capturing}
Table \ref{Qualitative_Results_Object_Capturing_Block} displays qualitative results for three of six objects. 
During Object Capturing, we took 105 images per object, similar to the perspectives shown in the first column of Table \ref{Qualitative_Results_Object_Capturing_Block}. Structure from Motion derived camera poses for all six objects of interest. The trajectories for these objects are similar to the trajectories displayed in the second row.

\subsection{Model Generation}
Object Capturing successfully processed all six objects, enabling Model Generation to create textured meshes for each object. Table \ref{Qualitative_Results_Model_Generation_Block} summarizes the qualitative performance of this process. In the beginning, Neural Rendering trained a NeuS-based Radiance Field for all objects. 
The first column of Table \ref{Qualitative_Results_Model_Generation_Block} shows views rendered from three of the six Radiance Fields. The remaining Radiance Fields produce similar results. All Radiance Fields generated novel views of high quality. Minor artifacts are present in challenging areas, such as the central wheel of the multimeter. This qualitative results indicate that the Radiance Fields learned accurate 3D representations, enabling photorealistic view synthesis.

\begin{table}[t]
    \centering
    \begin{tabular}{C C C}
        \hline
        % Column headers (centered)
        \textbf{Neural Rendering} & \textbf{Mesh Extraction} & \textbf{Mesh Refinement} \\
        \hline
        % Row 1 - Elephant
        \includegraphics[width=0.23\textwidth]{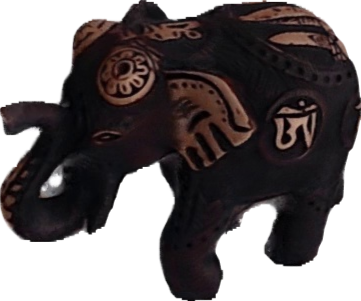} &
        \includegraphics[width=0.26\textwidth]{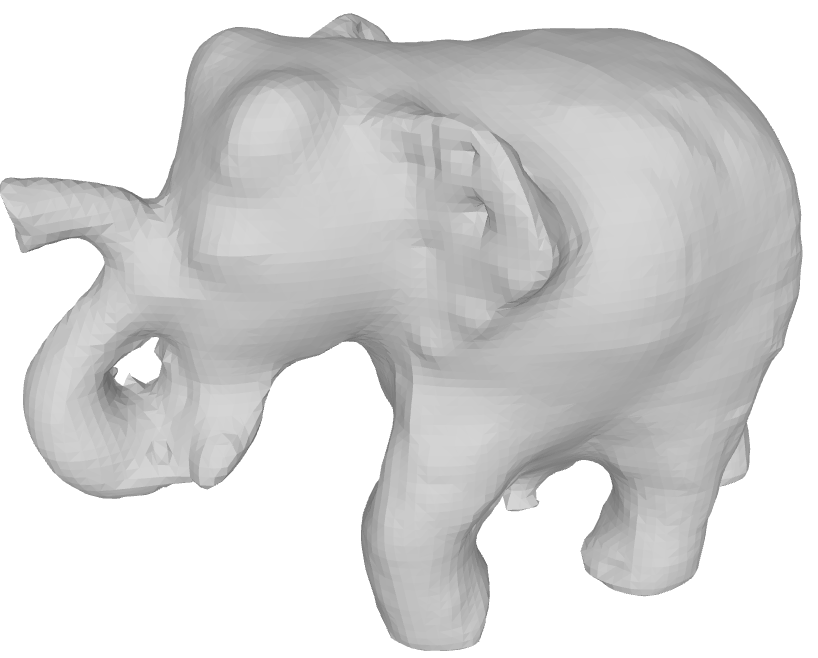} &
        \includegraphics[width=0.32\textwidth]{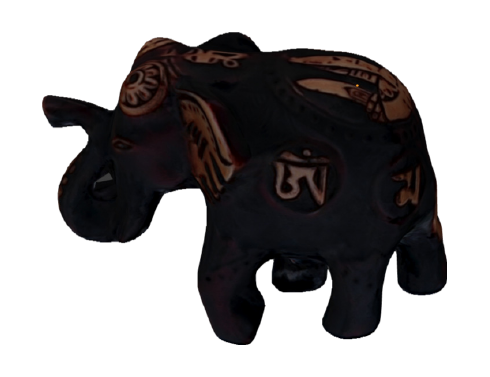} \\
        % Row 2 - Remote
        \includegraphics[width=0.25\textwidth]{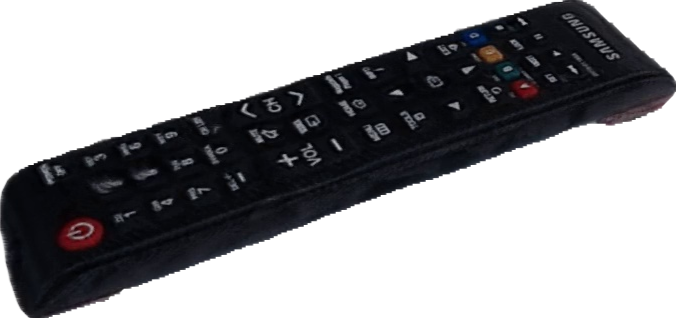} &
        \includegraphics[width=0.25\textwidth]{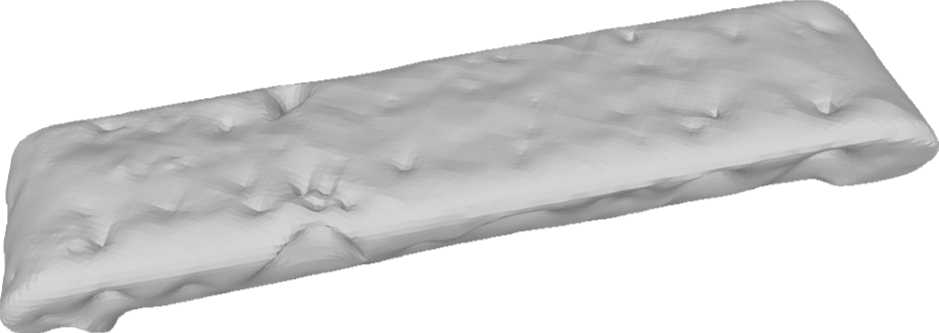} &
        \includegraphics[width=0.25\textwidth]{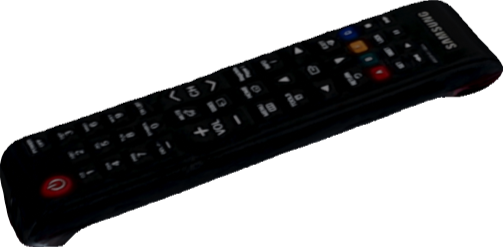} \\
        % Row 3 - Multimeter
        \includegraphics[width=0.25\textwidth]{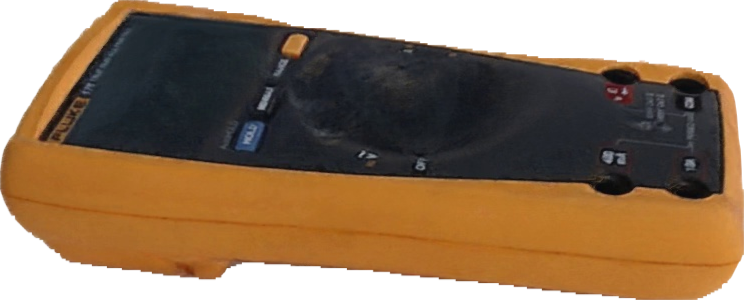} &
        \includegraphics[width=0.25\textwidth]{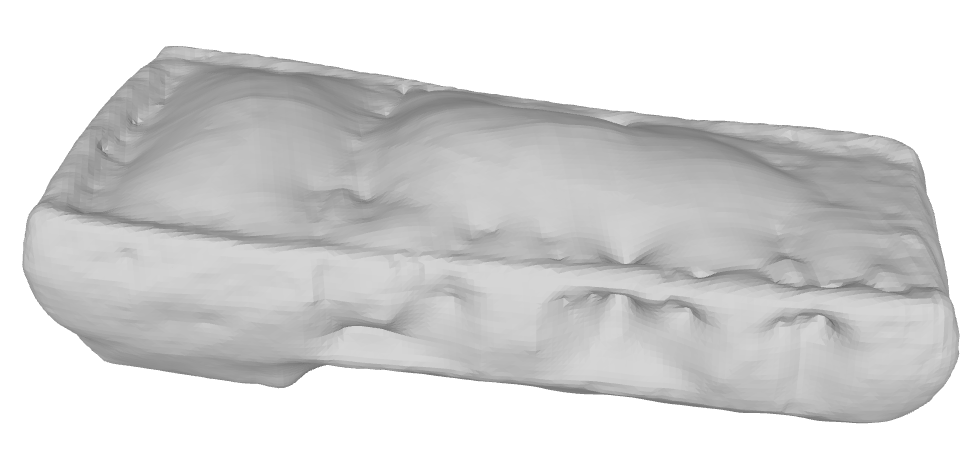} &
        \includegraphics[width=0.25\textwidth]{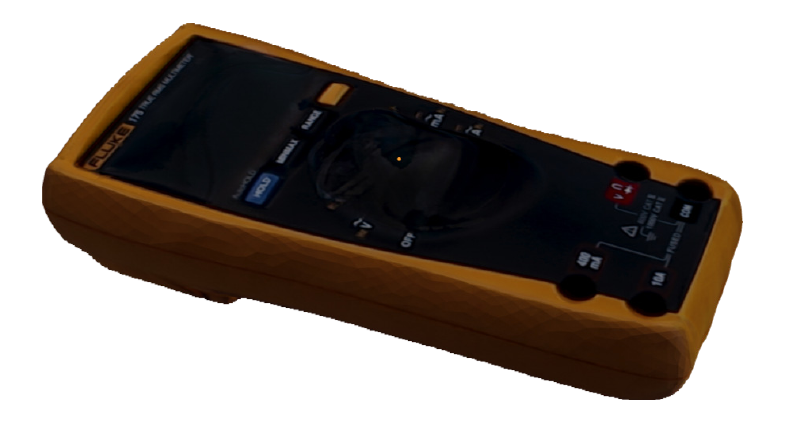} \\
        \hline
    \end{tabular}
    \caption{Model Generation for three target objects (elephant, remote, multimeter).}
    \label{Qualitative_Results_Model_Generation_Block}
\end{table}

\begin{table}[h!]
    \centering
    \begin{tabular}{lccc}
        \toprule
        & \textbf{PSNR [dB]} & \textbf{LPIPS} & \textbf{SSIM} \\ 
        \midrule
        
        Elephant    & 35.09 & 0.0105 & 0.994\\
        Cup         & 32.52 & 0.0349 & 0.983\\
        Baseball    & 38.61 & 0.0103 & 0.996\\
        Remote      & 32.72 & 0.0191 & 0.990\\
        Controller  & 32.70 & 0.0317 & 0.982 \\       
        Multimeter  & 34.58 & 0.0370 & 0.982 \\
        
        \midrule
        \textbf{Average} & 32.88 & 0.0239 & 0.9878 \\
        \bottomrule
    \end{tabular}
    \caption{Peak Signal to Noise Ratio (PSNR), Learned Perceptual Image Patch Similarity (LPIPS) and Structural Similarity Index Measure (SSIM) for the Radiance Fields, generated during the Neural Rendering Phase.}
    \label{tab:Performance_Neurel_Rendering}
\end{table}

Table \ref{tab:Performance_Neurel_Rendering} presents quantitative performance metrics of our Radiance Fields on real world test images. Other Radiance Field frameworks\cite{Nerf,InstantNGP,Delicate} achieve similar results on synthetic and real-world datasets of comparable complexity. Therefore, the quantitative results indicate successful 3D representation learning as well. 
Mesh Export extracted triangular meshes for all six target objects. Due to the NeuS-based Radiance Fields avoiding the unconstrained surface issue, the meshes exhibit relatively high surface quality. However, stronger rendering errors persist in the coarse mesh, for example in the central area of the multimeter. During Mesh Refinement, the coarse meshes were optimized, and the diffuse texture of the appearance model was exported for each object.\\

Mesh refinement improved vertex positions and mesh faces, correcting minor flaws in the coarse meshes, such as the holes in the elephant's ear. The textured meshes are displayed in the third row of Table~\ref{Qualitative_Results_Model_Generation_Block}.
The diffuse mesh textures appear darker than the rendered meshes from the previous phase because they do not account for the lighting conditions during capture. Overall, the meshes demonstrate high fidelity, comparable to the rendered views of the Radiance Fields.

\begin{figure}[t]
    \centering
    % First row
    \begin{subfigure}[b]{0.325\textwidth}  % First row, first column
        \centering
        \includegraphics[width=1\textwidth]{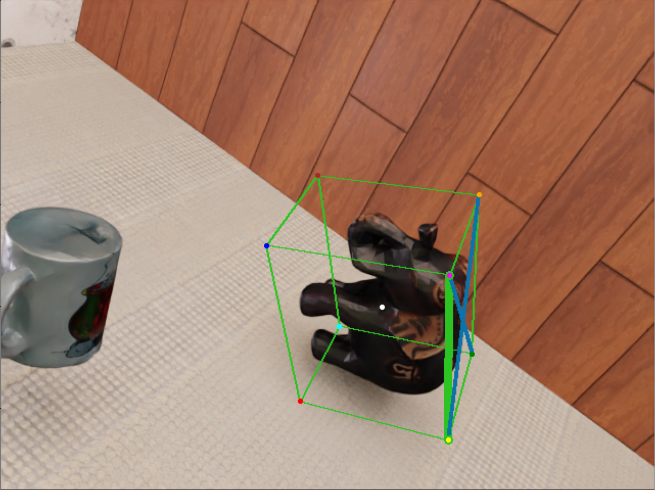}
        \caption{No occlusion}
        \label{fig:ele_no_occ}
    \end{subfigure}
    \hfill
    \begin{subfigure}[b]{0.325\textwidth}  % First row, second column
        \centering
        \includegraphics[width=1\textwidth]{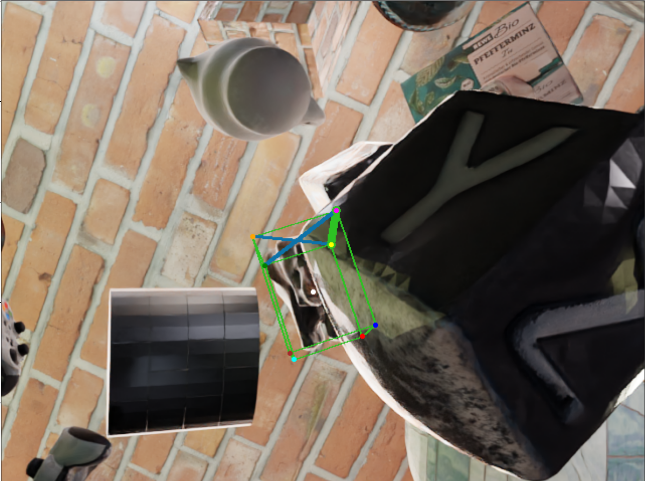}
        \caption{Moderate occlusion}
        \label{fig:ele_moderate_occ}
    \end{subfigure}
    \hfill
    \begin{subfigure}[b]{0.325\textwidth}  % First row, third column
        \centering
        \includegraphics[width=1\textwidth]{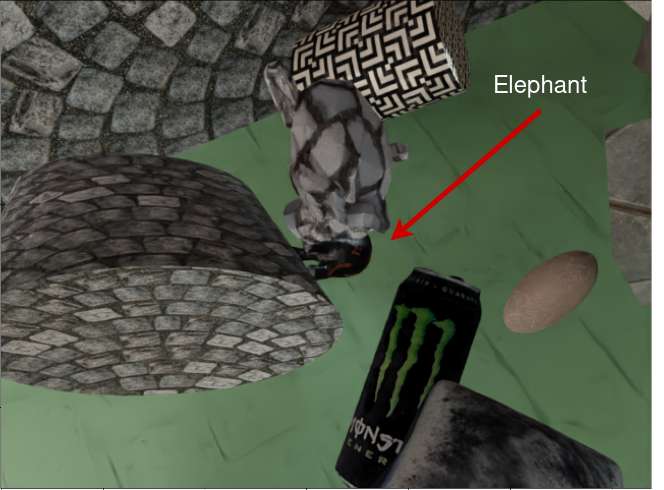}
        \caption{Strong occlusion}
        \label{fig:ele_strong_occ}
    \end{subfigure}

    % Second row
    \vskip\baselineskip % Add some vertical space between rows
    \begin{subfigure}[b]{0.325\textwidth}  % Second row, first column
        \centering
        \includegraphics[width=1\textwidth]{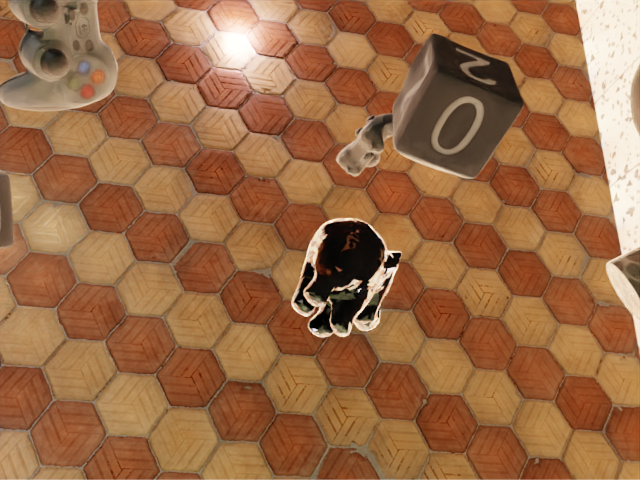}
        \caption{Camera on top}
        \label{fig:ele_additional_1}
    \end{subfigure}
    \hfill
    \begin{subfigure}[b]{0.325\textwidth}  % Second row, second column
        \centering
        \includegraphics[width=1\textwidth]{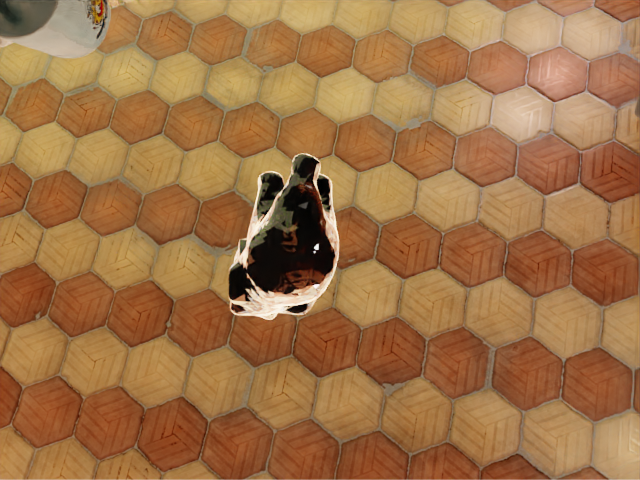}
        \caption{Camera behind}
        \label{fig:ele_additional_2}
    \end{subfigure}
    \hfill
    \begin{subfigure}[b]{0.325\textwidth}  % Second row, third column
        \centering
        \includegraphics[width=1\textwidth]{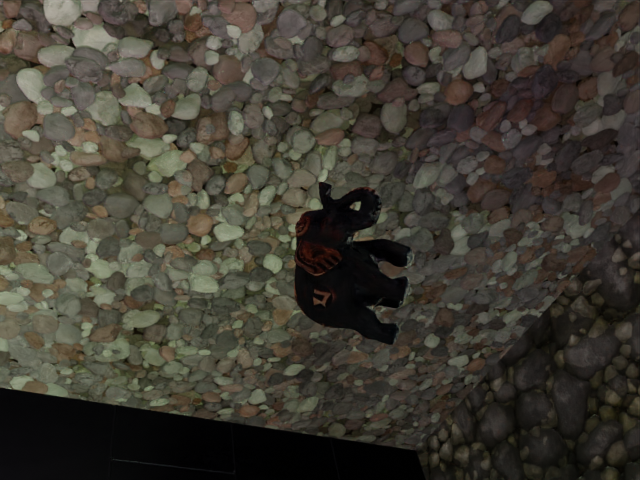}
        \caption{No direct light}
        \label{fig:ele_additional_3}
    \end{subfigure}

    \caption{Images, sampled from the elephant dataset, demonstrating, occlusion aware annotation (a, b, c) and photorealistic rendering (d, e, f).}
    \label{fig:ele_dataset}
\end{figure}

\subsection{Dataset Generation}
Our Dataset Generation process created a dataset for the elephant, the remote, and the multimeter. It aligned each mesh with Blender's internal coordinate system in the first step. Afterward, it applied Scene Composition  and Rendering + Annotation to create a synthetic dataset. 
Fig.~\ref{fig:ele_dataset} displays annotated and unannotated images of the elephant dataset, which were 
generated with this workflow. The first row of Fig.~\ref{fig:ele_dataset} demonstrates measurements to approach the content gap. The first row shows how domain randomization creates diverse backgrounds. Additionally, the second row demonstrates the photorealistic rendering capabilities of BlenderProc. 

\subsection{Pipeline Validation}
We trained and tested a pose estimation network on three generated datasets to verify our pipeline. The pose estimation model is based on the DOPE (Deep Object Pose Estimation)~\cite{DOPE} approach that predicts keypoint belief maps and their corresponding affinity maps. 
DOPE has to be trained per object, therefore we trained 3 different models. Each DOPE network was trained for 100 epochs, with a learning rate of 0.0001 and a batch size of 64. 

To verify our pipeline in practical scenarios, we tested the DOPE models for two scenarios: tabletop pose estimation and in-hand pose estimation. Since we generated datasets for custom objects, no public datasets are available to apply these scenarios. Furthermore, a real world test set is not attainable, without significant effort, due to the complex annotation process. To still assess the model’s performance, we recorded test sequences for both tasks and performed a qualitative evaluation.

\begin{figure}[t]
    \centering
    % First column
    \begin{subfigure}[b]{0.325\textwidth}  % First column, first row
        \centering
        \includegraphics[width=1\textwidth]{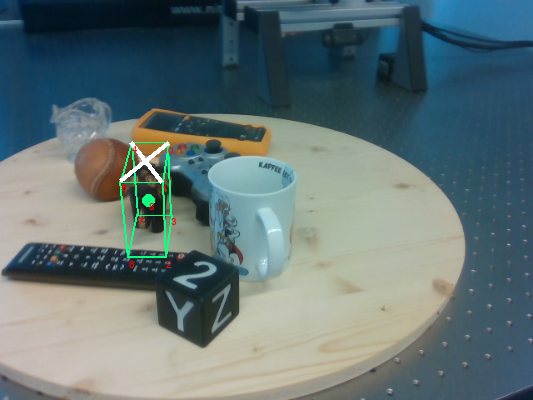}
        \label{fig:ele_good}
    \end{subfigure}
    \hfill
    \begin{subfigure}[b]{0.325\textwidth}  % Second column, first row
        \centering
        \includegraphics[width=1\textwidth]{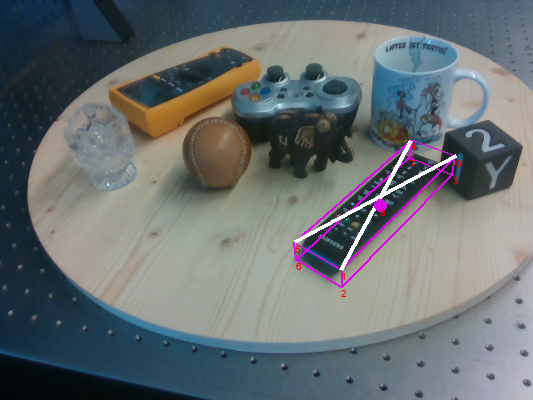}
        \label{fig:multi_tabletop_1}
    \end{subfigure}
    \hfill
    \begin{subfigure}[b]{0.325\textwidth}  % Third column, first row
        \centering
        \includegraphics[width=1\textwidth]{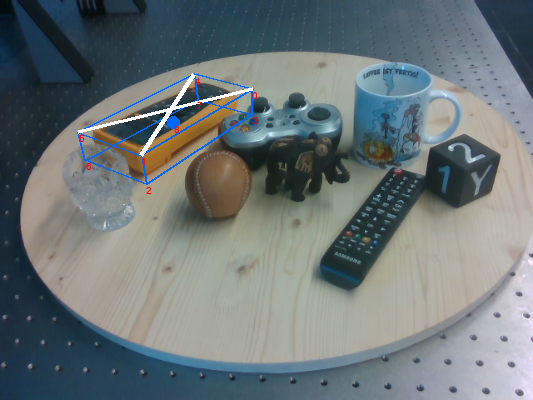}
        \label{fig:multi_tabletop_1}
    \end{subfigure}

    % Second row
    %\vskip\baselineskip
    \begin{subfigure}[b]{0.325\textwidth}  % First column, second row
        \centering
        \includegraphics[width=1\textwidth]{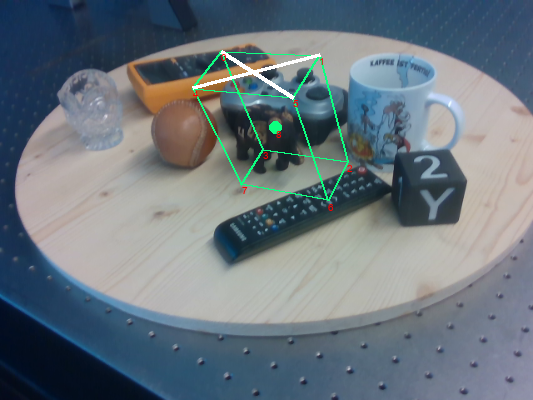}
        \label{fig:ele_okay}
    \end{subfigure}
    \hfill
    \begin{subfigure}[b]{0.325\textwidth}  % Second column, second row
        \centering
        \includegraphics[width=1\textwidth]{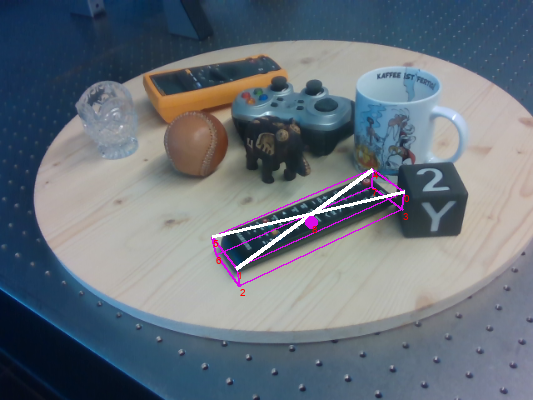}
        \label{fig:multi_tabletop_2}
    \end{subfigure}
    \hfill
    \begin{subfigure}[b]{0.325\textwidth}  % Third column, second row
        \centering
        \includegraphics[width=1\textwidth]{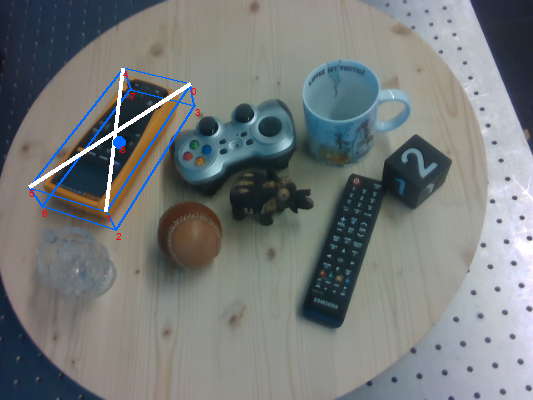}
        \label{fig:multi_tabletop_2}
    \end{subfigure}

    \caption{Tabletop pose estimation with elephant (first column), remote (second column), and multimeter (third column) as target objects.}
    \label{fig:tabletop}
\end{figure}

\subsubsection{Tabletop Pose Estimation}
\begin{figure}[t]
    \centering
    \begin{subfigure}[b]{0.325\textwidth}  % First column
        \centering
        \includegraphics[width=1\textwidth]{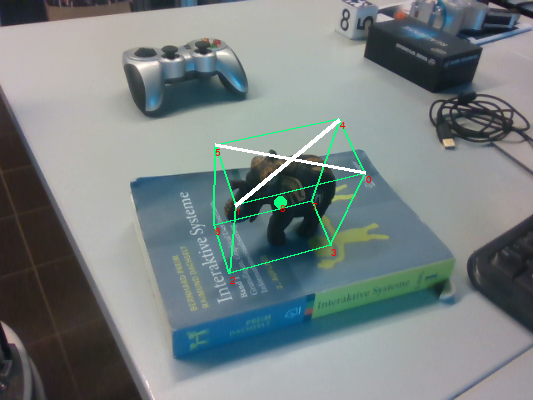}
        \label{elephant_table_simp_1}
    \end{subfigure}
    \hfill
    \begin{subfigure}[b]{0.325\textwidth}  % Second column
        \centering
        \includegraphics[width=1\textwidth]{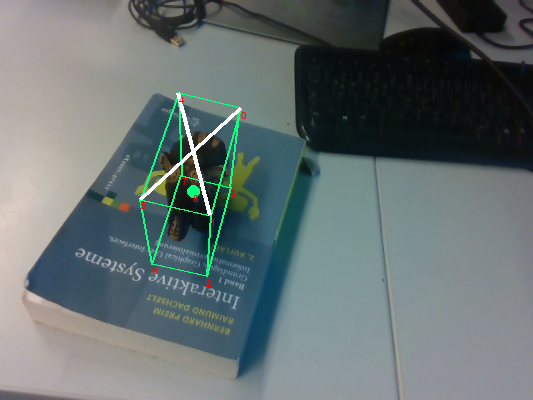}
        \label{elephant_table_simp_2}
    \end{subfigure}
    \hfill
    \begin{subfigure}[b]{0.325\textwidth}  % Third column
        \centering
        \includegraphics[width=1\textwidth]{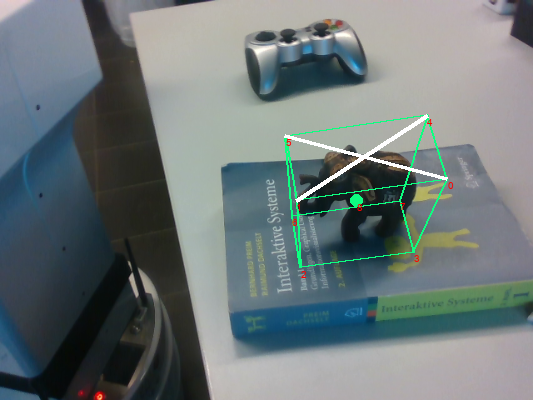}
        \label{elephant_table_simp_3}
    \end{subfigure}
    
    \caption{Table-top pose estimation for the elephant in a simpler scenario.}
    \label{fig:ele_tabletop_simple}
\end{figure}

To test our networks on tabletop pose estimation, we created a scene which contains the target objects and some distractors.
All three networks were tested on this sequence, qualitative results of this process are displayed in Fig. \ref{fig:tabletop}.
The results for the multimeter and the remote indicate successful pose estimation with correct bounding box locations. However, the performance of the
DOPE network trained on the elephant dataset was notably worse.
Unlike the other two objects, the elephant was evaluated against a complex background. This complexity may have contributed to the detection challenges. To address this issue, we created a test sequence, which solely focuses on the elephant was Fig. \ref{fig:ele_tabletop_simple} shows four frames of this sequence. The DOPE model performed significantly better, showing strong performance.

\begin{figure}[ht]
    \centering
    % First row
    \begin{subfigure}[b]{0.325\textwidth}  % First row, first column
        \centering
        \includegraphics[width=1\textwidth]{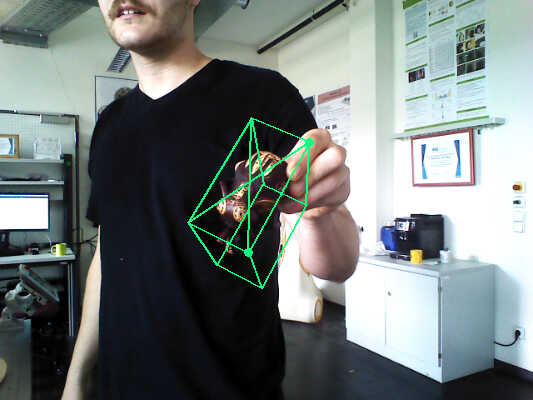}
        \label{fig:ele_good}
    \end{subfigure}
    \hfill
    \begin{subfigure}[b]{0.325\textwidth}  % First row, second column
        \centering
        \includegraphics[width=1\textwidth]{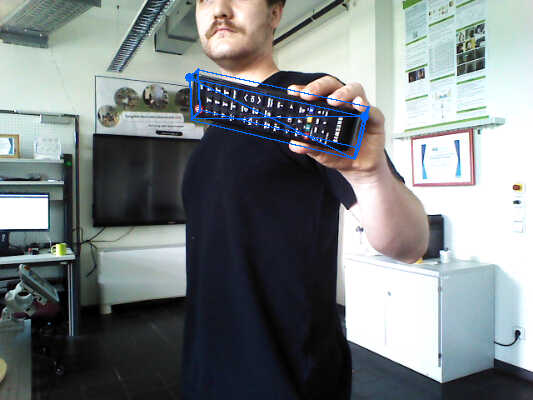}
        \label{fig:remote_inhand_1}
    \end{subfigure}
    \hfill
    \begin{subfigure}[b]{0.325\textwidth}  % First row, third column
        \centering
        \includegraphics[width=1\textwidth]{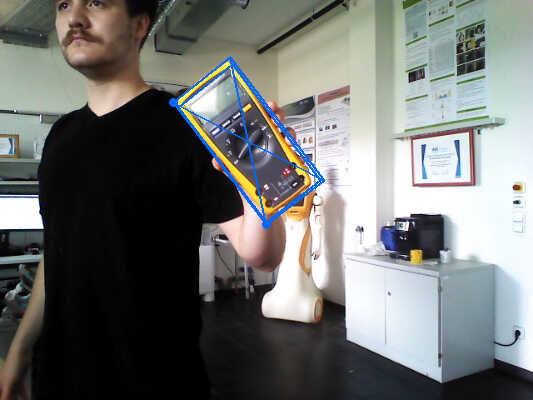}
        \label{fig:multi_inhand_1}
    \end{subfigure}
    
    % Second row
    %\vskip\baselineskip % Add some vertical space between rows
    \begin{subfigure}[b]{0.325\textwidth}  % Second row, first column
        \centering
        \includegraphics[width=1\textwidth]{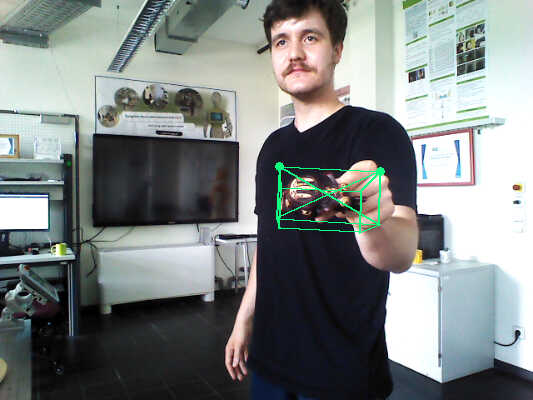}
        \label{fig:ele_okay}
    \end{subfigure}
    \hfill
    \begin{subfigure}[b]{0.325\textwidth}  % Second row, second column
        \centering
        \includegraphics[width=1\textwidth]{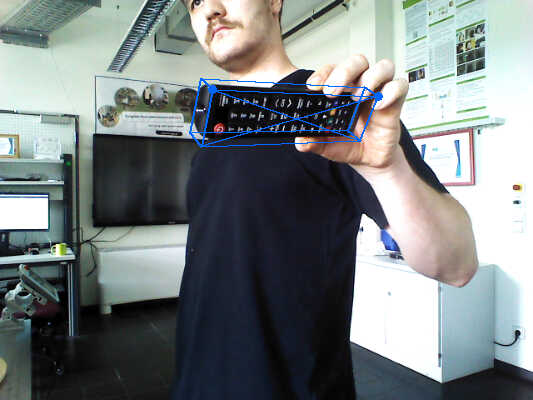}
        \label{fig:remote_inhand_2}
    \end{subfigure}
    \hfill
    \begin{subfigure}[b]{0.325\textwidth}  % Second row, third column
        \centering
        \includegraphics[width=1\textwidth]{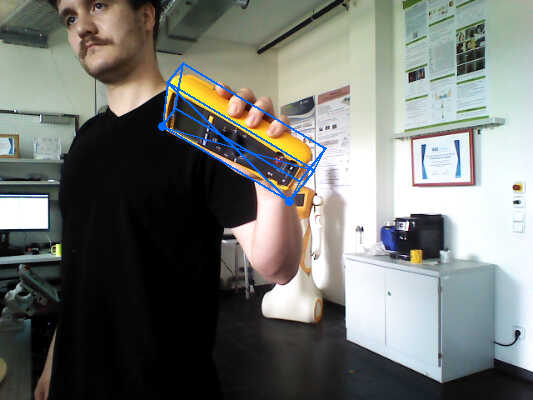}
        \label{fig:multi_inhand_2}
    \end{subfigure}

    \caption{Qualitative results during in-hand detection for the three target objects.}
    \label{fig:in_hand}
\end{figure}

\subsubsection{In-Hand Pose Estimation}
We recorded a human, holding the target object in different angles, in front of a camera, to derive the test sequences for in-hand pose estimation. The six frames, depicted in Fig.~\ref{fig:in_hand} showcases the performance of our the three models. 
The multimeter and remote networks achieved a performance similar to that of the tabletop scenario. The elephant network produced more accurate bounding boxes compared to the previous scenario. However, the performance of the elephant network was still worse than the performance of the remaining two networks.

\subsection{Limitations}
Our pipeline has some limitations related to the general concept and specific pipeline blocks. The pipeline is limited to the front side of the target objects, it cannot reconstruct the back side of objects, since these parts are not present during Radiance Field training. Subsequently, the back side can neither be meshed nor detected.
Furthermore, the generated meshes show high fidelity but contain small errors in ambiguous regions, like the central regions of the multimeter. These errors are introduced during Neural Rendering and inherited in the subsequent phases. This limitation might restrict our pipeline to objects of lower complexity.

\section{Conclusion and Future Work}
In this work, we proposed a pipeline that utilizes Radiance Fields to automatically generate datasets for the 3D pose estimation of arbitrary objects. In contrast to previous works, our pipeline does not require 3D models of the object of interest to generate datasets. Our experiments demonstrated that our pipeline can generate high-quality textured meshes for objects of varying complexity. We also showed that the pipeline is sufficient to train 3D pose estimation models that work in practical scenarios. Our pipeline only involves two manual steps: placing the target objects on the turntable and aligning the model with Blender's internal coordinate system. Based on these results, we conclude that Radiance Fields can provide 3D models which are sufficient to generate high-quality synthetic datasets in an automated fashion.

Future works will address the conceptual weakness of our pipeline by expanding the model generation to full object reconstruction. Generating two models per object and fusing them is an approach we might pursue in upcoming publications to archive this. 
Expanding the pipeline to stereo 3D datasets is another promising research direction, we might approach in future works. 
This could be realized by creating and leveraging stereo images to improve object detection accuracy and enable depth perception. Furthermore, we will expand our qualitative evaluation with a quantitative evaluation. We will apply our pipeline to a benchmark dataset, which contains 3D models and a real world test set, is a way to approach this goal.

\begin{credits}
\subsubsection{\ackname} This work is funded and supported by the Federal Ministry of Education and Research of Germany (BMBF) (AutoKoWAT-3DMAt under grant No. 13N16336), German Research Foundation (DFG) (SEMIAC under grant number No. 502483052) and by the European Regional Development Fund (ERDF) (ENABLING under grant No. ZS/2023/12/182056) and (RoboLab under grant No. ZS/2023/12/182065).

\subsubsection{\discintname}
The authors have no competing interests to declare that are
relevant to the content of this article. 
\end{credits}


\bibliographystyle{unsrt}	% only numbers
\bibliography{content}

\end{document}